\documentclass[runningheads]{llncs}
\usepackage{graphicx}
\usepackage{tikz}
\usepackage{multirow}
\usepackage{enumitem}
\usepackage{amssymb}
\usepackage{array} 
\usepackage{mathrsfs} 
\usepackage{algorithm}
\usepackage{amsmath}
\usepackage{bm}
\usepackage{algpseudocode}
\usepackage{tablefootnote}
\usepackage{hyperref}

\newcolumntype{P}[1]{>{\centering\arraybackslash}p{#1}}
\begin{document}
\title{A novel structured argumentation framework for improved explainability of classification tasks}

\titlerunning{A novel structured argumentation framework...}
%
\author{Lucas Rizzo\inst{1}\orcidID{0000-0001-9805-5306} and Luca Longo\inst{1}\orcidID{0000-0002-2718-5426}}
%
\authorrunning{L. Rizzo and L. Longo}
%
\institute{Technological University Dublin, Dublin, Ireland\\
\email{lucas.rizzo@tudublin.ie}}
%
\maketitle              
\begin{abstract}

This paper presents a novel framework for structured argumentation, named extend argumentative decision graph ($xADG$). It is an extension of argumentative decision graphs \cite{dondio2021towards} built upon Dung's abstract argumentation graphs. The $xADG$ framework allows for arguments to use boolean logic operators and multiple premises (supports) within their internal structure, resulting in more concise argumentation graphs that may be easier for users to understand. The study presents a methodology for construction of $xADGs$ and evaluates their size and predictive capacity for classification tasks of varying magnitudes. Resulting $xADGs$ achieved strong  (balanced) accuracy, which was accomplished through an input decision tree, while also reducing the average number of supports needed to reach a conclusion. The results further indicated that it is possible to construct plausibly understandable $xADGs$ that outperform other techniques for building $ADGs$ in terms of predictive capacity and overall size. In summary, the study suggests that $xADG$ represents a promising framework to developing more concise argumentative models that can be used for classification tasks and knowledge discovery, acquisition, and refinement.

\end{abstract}

\section{Introduction}

Several works have employed argumentation and machine learning (ML) in different fashions \cite{cocarascu2016argumentation}. Their choice
of argumentation framework and purpose of employing argumentation differ, for example, by attempting to improve the performance or the explanatory power of ML models. This paper focus on the problem of explainability. In other words, it proposes an argumentation framework that could be used to improve the understandability of data-driven models. In particular,
the emphasis is on decision trees (DT) and on defining argumentative models of equivalent inferential capability but that could be perceived as more understandable. The concept of understandability, as well as the associated notions related to explainability, have a plethora of definitions and can be measured in different way \cite{VILONE202189}. Succinctly, understandability could be linked to the capacity of a model to be understandable from its user point of view. The qualitative analysis of understandability is not performed in this paper. Instead, it provides a quantitative analysis that could indicate a better perceived understandability. 
It is often inferred that argumentative models are inherently  transparent and comprehensible, as they may employ natural language terms and attempt to follow the way humans reason. However, even apparently transparent systems can become convoluted when dealing with large problems. For instance, it is reasonable to assume that an argumentative model with a few arguments is easier to understand than a DT with a plethora of nodes and edges. Hence, it seems worthy to pursue frameworks and automated alternatives for constructing accurate and understandable argumentative models in different contexts.

In summary, this paper presents a novel framework for structured argumentation named extend argumentative decision graph ($xADG$), which enables arguments to use boolean logic operators and multiple supports within their internal structure. The proposed framework is an extension of the argumentative decision graph ($ADG$) developed by \cite{dondio2021towards}, which was itself an extension of Dung's abstract argumentation graphs \cite{DUNG1995321}. The construction of inferential models using $xADG$ is proposed to be made from a given DT. This way, while both the DT and derived argumentative model are guaranteed to maintain the same inferential capability, they may differ in terms of size and comprehensibility. The expectation is that a smaller model will be easier for human reasoners to understand and extend. A preliminary analysis is proposed to investigate whether reasonably smaller structures, in terms of number of arguments/attacks and amount of argument supports, can be achieved for classification tasks of different sizes in the UCI machine learning repository \cite{uci-ml}. 

The remainder of this paper is organised as follows. In Section 2, the basics of abstract argumentation semantics and argumentative decision graphs are reviewed. Similar works that have attempted to solve similar problems are also described. Section 3 introduces the formal definition of $xADG$ and describes the experiment proposed to build inferential models using it. Results and discussion are presented in Section 4. Finally, Section 5 concludes the study and gives a number of indications for future work.

\section{Literature Review}

When engaging in computational argumentation, the first step is typically to generate a collection of arguments, which can be abstract (symbolic) or structured in various ways. From there, one can establish relationships of attack or support between them, resulting in a network of interrelated arguments known as an argumentation graph. The literature on creating argumentation structures and determining their acceptance in a given context is vast \cite{baroni2011introduction}.
However, there appears to be a lack of research focused on automatically generating arguments and interactions from data, a critical component for advancing the use and deployment of argumentation systems. The alternative - manual knowledge acquisition from domain experts - is usually prohibitively time-consuming \cite{RIZZO2023537}. A few works have performed experiments in this line. For example,
the authors in \cite{thimm2017towards} propose the extraction of contradicting rules based on frequent patterns found in the data, followed by their use in structured argumentation, such as ASPIC+ \cite{modgil2014aspic}. Preliminary results show that this two-step approach performed well for instances that could be classified, but returned a large amount of undecided records. In \cite{cocarascu2020data}, the authors put forward a method for extraction of argumentation frameworks for individual unlabelled instances in a dataset, based on other labelled instances. The approach demonstrates competitive predictive capacity for different types of data, but does not analysis the explainability of results in full. In \cite{noor2020bayesian}, argument attacks are extracted from data-driven proxy indicators through which the probability that a set of arguments is accepted or rejected can be inferred. Experiments are conducted using synthetic data and showing the feasibility of the approach.

Some other works propose the creation of complete or approximate mappings from black-box models to argumentation frameworks. For example, in \cite{bistarelli2022argumentative}, argumentation is employed  as a means of explanation for machine learning outcomes, providing an argumentative interpretation of both the training process and outcomes. In \cite{rago2020argumentation} the authors provide a recommender system by extracting argumentation explanations in the form of bipolar argumentation frameworks. In \cite{collins2019towards} 
an approach for extracting an argumentation framework from a planning model is given. It is designed in such way to allow its user to query it for explanations about its outputs.
Hence, such mappings are usually employed to generate comprehensible explanations or  are only assumed, and not investigated, to be more understandable given the inherently transparency of argumentative systems. In contrast, in this work a novel structured argumentation framework is proposed, which is concerned with both predictive capacity and model understandability. A transparent and more concise model could enhance its user capacity of knowledge discovery and refinement.
The next subsections introduce the necessary concepts for the definition of the proposed framework.



\subsection{Abstract argumentation semantics}

In this paper, it is important to define the notions of reinstatement and conflictfreeness, as well as the most common Dung semantics \cite{DUNG1995321}, such as grounded and preferred. The definitions follow from the the works in \cite{caminada2009logical,caminada2006issue}.

\begin{definition}
Let $\langle Ar , \mathscr{R}\rangle$ be an Argumentation Framework (AF), $a, b \in Ar$ and $args \subseteq Ar$. The following shorthand notations are employed:

\begin{itemize}
 \item[\textbullet] $a^+$ as $\{b\,|\,(a,b) \in \mathscr{R}\}$.
 \item[\textbullet] $args^+$ as $\{a\,|\,(a,b) \in \mathscr{R}$ for some $a \in args\}$.
 \item[\textbullet] $a^-$ as $\{b\,|\,(b,a) \in \mathscr{R}\}$.
 \item[\textbullet] $args^-$ as $\{b\,|\,(b,a) \in \mathscr{R}$ for some $a \in args\}$.
\end{itemize}
\end{definition}

$a^+$ indicates the arguments attacked by $a$, while $a^-$ indicates the arguments attacking $a$. $args^+$ indicates the set of arguments attacked by $args^+$, while $args^-$ indicates the set of arguments attacking $args^-$.

\begin{definition}
Let $\langle Ar , \mathscr{R}\rangle$ be an AF and $args \subseteq Ar$. $args$ is \textbf{conflict-free} iff $args \cap args^+ = \emptyset$.
\end{definition}

\begin{definition}
Let $\langle Ar , \mathscr{R}\rangle$ be an AF, $A \in Ar$ and $args \subseteq Ar$. $args$ \textbf{defends} an argument $A$ iff $A^- \subseteq args^+$.
\end{definition}

\begin{definition}
Let $\langle Arg, \mathscr{R} \rangle$ be an AF and $Lab : Arg \rightarrow \{in, out,$ $undec\}$ be a labelling function. $Lab$ is a 
\textbf{reinstatement labelling} iff it satisfies:

\begin{itemize}
 \item $\forall A \in Ar : (Lab(A) = out \equiv \exists B \in Ar : (B$ \textit{defends} $A \wedge Lab(B) = in))$ and
 \item $\forall A \in Ar : (Lab(A) = in \equiv \forall B \in Ar : (B$ \textit{defends} $A \supset Lab(B) = out))$ 
\end{itemize}
\end{definition}

\begin{definition}\label{def:dungssemantics}
Let $Args$ be a conflict-free set of arguments, $F : 2^{Args} \rightarrow 2^{Args}$ a function such that $F(Args) = \{A\,|\,A$ is defended by $Args\}$ and
$Lab : Args \rightarrow \{in, out, undec\}$ a reinstatement labelling function. Also consider $in(Lab)$ short for $\{A \in Args\,|\,Lab(A) = in\}$, 
$out(Lab)$ short for $\{A \in Args\,|\,Lab(A) = out\}$ and $undec(Lab)$ short for $\{A \in Args\,|\,Lab(A) = undec\}$.

\begin{itemize}
 \item[\textbullet] $Args$ is \textbf{admissible} if $Args \subseteq F(Args)$.
 \item[\textbullet] $Args$ is a \textbf{complete} extension if $Args = F(Args)$.
 \item[\textbullet] $in(Lab)$ is a \textbf{grounded} extension if $undec(Lab)$ is maximal, or $in(Lab)$ is minimal, or $out(Lab)$ is minimal.
 \item[\textbullet] $in(Lab)$ is a \textbf{preferred} extension if $in(Lab)$ is maximal or $out(Lab)$ is maximal.
\end{itemize}
\end{definition}

The use of argumentation semantics, such as grounded and preferred, can help us examine the justification status of arguments. Essentially, an argument
is justified if it can somehow withstand its attackers.
Thus, argumentation semantics provide a perspective
that one can take, when deciding on the set of accepted, rejected, and undecided arguments. The extension (set of acceptable arguments) can defend itself and remain internally coherent, even if
someone disagrees with its viewpoint \cite{wu2010labelling}. The grounded semantics is considered more sceptical, as it takes fewer committed choices and always offers a single extension. For the purposes of this paper, using the grounded semantics is adequate for deriving the inferences from the proposed classification tasks.

\subsection{Argumentative decision graphs}

Formally, the following shorthand notations are employed:

\begin{itemize}
    \item[\textbullet] A dataset $\mathscr{D}$ is represented by a $N \times M$ matrix-like data structure. Each row is called an \textit{instance} of the dataset and each column is called a \textit{feature}.
    \item[\textbullet] The predicates of the form $f_i(v)$ mean \textit{``the feature $i$ has the value $v$"}.
     \item[\textbullet] The set of all values available for $f_i$ is $\mathscr{V}_{f_i}$.
    \item[\textbullet] One of the features $f_i$ is called the target variable and it is denoted with $y$.
    \item[\textbullet] The predicates involving the target variable have the form $y(v)$.
    \item[\textbullet] $\mathscr{P}_f$ is the set of predicates regarding the other features $\{f_1,\ldots,f_m\}$.
    \item[\textbullet] $\mathscr{P}_y$ is the set of predicates regarding the target variable $y$.
    \item[\textbullet] The set of all the predicates is called $\mathscr{P}_f \cup \mathscr{P}_y$.
\end{itemize}

An $ADG$ is proposed in \cite{dondio2021towards} and is defined as following:

\begin{definition}\label{def:adg}
An argumentative decision graph \textbf{ADG}, is an argumentation framework $AF =
\langle Ar, \mathscr{R} \rangle$, where each $a \in Ar$ is defined as $a = \langle$$\phi$, $\theta$$\rangle$, $\phi \in \mathscr{P}_f, \theta \in \mathscr{P}_y \cup \varnothing$, and $\mathscr{R} \subseteq Ar \times Ar$.
\end{definition}

An $ADG$ is a type of abstract argumentation graph that enhances Dung's original model. $ADGs$ have a rule-based structure where each argument has a single premise (support) and a conclusion. The support consists of a feature and a corresponding value, while the conclusion represents a value for the target variable. Some arguments may not have a conclusion, and instead, they interact with other predictive arguments.
The notion of well-formed $ADGs$ is also introduced by \cite{dondio2021towards}:

\begin{definition}\label{def:wellformed}
Given an $ADG = \langle Ar, \mathscr{R} \rangle$, the $ADG$ is \textbf{well-formed} iff $\forall a_1, a_2 \in Ar$ with $a_1 = \langle \phi_1, \theta_1 \rangle$, $a_2 = \langle \phi_2, \theta_2 \rangle$ it holds that:

\begin{enumerate}
    \item if $\phi_1 = f_i(v_1)$ and $\phi_2 = f_i(v_2)$ then $(a_1, a_2) \notin \mathscr{R} \land (a_2, a_1) \notin \mathscr{R}$.
    \item if $\theta_1 = \theta_2 \neq \varnothing$ then $(a_1, a_2) \notin \mathscr{R} \land (a_2, a_1) \notin \mathscr{R}$
    \item if $a_1 = \langle f_1(v_x), y(v_1) \rangle$, $a_2 = \langle f_2(v_y), y(v_2) \rangle$ and $v_1 \neq v_2 \land f_1 \neq f_2$ then $(a_1, a_2) \in \mathscr{R} \lor (a_2, a_1) \in \mathscr{R}$
\end{enumerate}
\end{definition}

Informally, these constraints guarantee that: 1) there is no attack between two arguments whose supports contain the same feature; 2) there is no attack between two arguments whose conclusions are the same and
not empty; and 3) if two arguments have mutually exclusive conclusions and they use different features in 
their supports, there must be an attack between them. A well-formed $ADG$ can be extracted from a DT following the steps in Algorithm \ref{alg:wbadg}. Intuitively, each terminal node in the DT will generate a predictive argument in the $ADG$, while non terminal nodes will generate non predictive arguments. Lastly, arguments with different features and conclusions that are in disjoint paths to distinct terminal nodes will generate attacks.
When exploited by an extension-based semantics, such as grounded, the resulting $ADG$ will produce the same set of inferences as the DT.

\renewcommand{\algorithmicrequire}{\textbf{Input:}}
\renewcommand{\algorithmicensure}{\textbf{Output:}}
\algnewcommand\algorithmicforeach{\textbf{for each}}
\algdef{S}[FOR]{ForEach}[1]{\algorithmicforeach\ #1\ \algorithmicdo}
\begin{algorithm}[h]
\caption{\footnotesize Algorithm for building a well-formed ADG from a DT \cite{dondio2021towards}}\label{alg:wbadg}
\footnotesize
\begin{algorithmic}[1]
\Require A DT
\Ensure A well-formed $ADG = \langle Ar, \mathscr{R}\rangle$
\State $Ar \gets \emptyset$, $\mathscr{R} \gets \emptyset$
\ForEach {directed link $l$ from a node $N$ to a terminal node $M$}
\State $Ar \gets Ar \cup \langle f_i(v_n), y_m\rangle$ 
\Comment{\parbox[t]{.6\linewidth}{$f_i$ is the variable tested at node $N$.}}
\Statex \Comment{\parbox[t]{.6\linewidth}{$v_n$ is the value of $f_i$ identifying the link $l$ connecting $N$ to the terminal node $M$.}}
\Statex \Comment{\parbox[t]{.6\linewidth}{$y_m$ is the value of the target variable predicted at node M.}}
\EndFor

\ForEach {directed link $l$ from a node $N$ to a non terminal node $M$}
\State $Ar \gets Ar \cup \langle f_i(v_n), \emptyset\rangle$ 
\Comment{\parbox[t]{.6\linewidth}{$f_i$ is the variable tested at node $N$.}}
\Statex \Comment{\parbox[t]{.6\linewidth}{$v_n$ is the value of $f_i$ identifying the link $l$ connecting $N$ to the terminal node $M$.}}
\EndFor

\ForEach {$a \in Ar$ associated to a link $l_a$ from node $N$ to $M$}
\ForEach {$b \in Ar$ associated to a link $l_b$}
\If{$l_b$ is in path $p$ connecting $N$ to a terminal node so that $l_a \in p$}
\State \textbf{continue} \Comment{\parbox[t]{.7\linewidth}{if $l_a$ and $l_b$ are in a same path from $N$ to a terminal node, then there is no attack}}
\EndIf
\If{$a$ and $b$ features are different or conclusions are different}
\State $\mathscr{R} \gets \mathscr{R} \cup (a,b)$
\EndIf
\EndFor
\EndFor
\\
\Return $ADG = \langle Ar, \mathscr{R}\rangle$
\end{algorithmic}
\end{algorithm}

    
    

\section{Design and methodology}

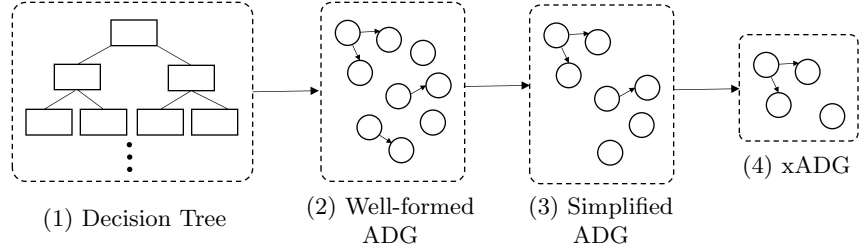
\begin{figure}[h]
    \centering
\begin{tikzpicture}
\node[inner sep=0pt] (diagram) at (0,0)
    {\includegraphics[width=0.95\textwidth]{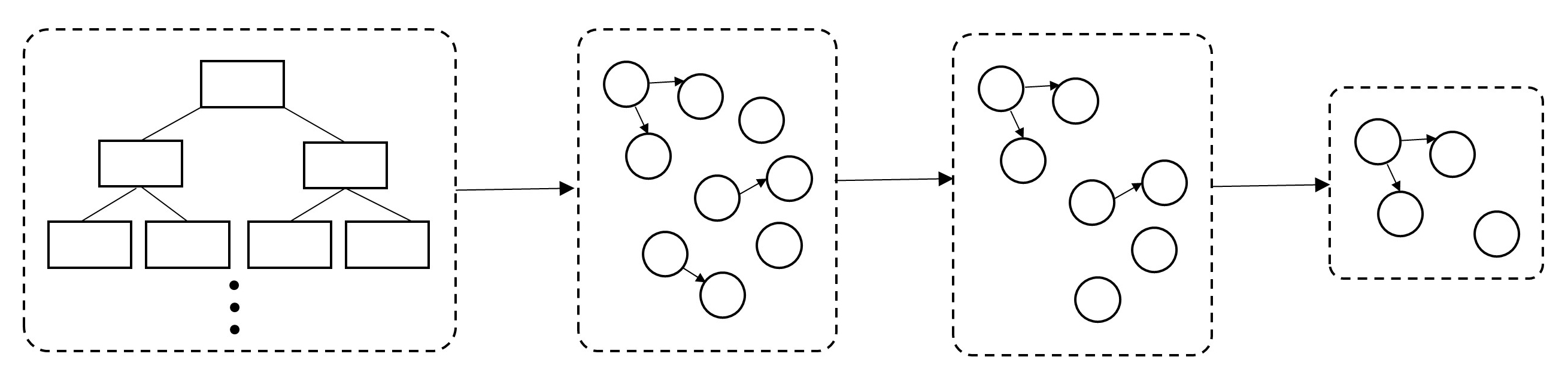}};
 \node[inner sep=0pt] (dt) at (-4,-1.7) {(1) Decision Tree};
 \node[inner sep=0pt, align=center] (dt) at (-0.6,-1.7) {(2) Well-formed \\ ADG};
 \node[inner sep=0pt, align=center] (dt) at (2.2,-1.7) {(3) Simplified \\ ADG};
 \node[inner sep=0pt] (dt) at (4.8,-1.0) {(4) xADG};
\end{tikzpicture}
\caption{Design of the proposed process to build a xADG from a DT while maintaining the same inferential capability. The goal is to achieve a smaller structure, that can be easier to understand and expand by human reasoners.}
    \label{fig:experiment}
\end{figure}

$ADGs$ are first presented \cite{dondio2021towards} as graph-like models that have the ability to perform classification tasks. In turn, this paper proposes a new model, called $xADG$, which extends $ADGs$ by allowing arguments to employ boolean logic operators and multiple supports. Later in this section, the precise definition of a $xADG$ is provided. Subsequently, the construction of $xADGs$ is proposed to be made from well-formed $ADGs$ built from DTs. A simplification procedure is proposed to eliminate any superfluous arguments that may arise from redundant branches within the tree. Afterwards, a set of modifications is introduced to build a $xADG$ from the resulting simplified $ADG$. These modifications will allow the removal of arguments and attacks while maintaining the same inferential capability, but incorporating the higher complexity permitted into the argument supports. Other methods have been proposed for the construction of well-formed $ADGs$ that do not rely on DTs or any other data-driven approaches \cite{dondio2021towards}.
However, these methods have yet to demonstrate superior predictive capacity or model size compared to data-driven classifiers such as DTs. Figure \ref{fig:experiment} depicts the process proposed for achieving $xADGs$ from DTs. The DT can be built by any standard algorithm, such as C4.5 or CART. In turn, a well-formed $ADG$ can be derived as detailed in Algorithm \ref{alg:wbadg}. The remaining of this section details the subsequent steps: the simplification of $ADGs$ and possible modifications to derive $xADGs$.

\subsection{ADG simplification}

When considering $ADGs$ built from DTs, it is important to note that a DT may contain two different branches that originate from the same starting point and ultimately lead to the same outcome. Hence, all the nodes in these two paths can be removed, while the node in the starting point can be updated in order to draw the same conclusion. 
Algorithm \ref{alg:wbadg} does not take into account such cases. Therefore, Algorithm \ref{alg:wbadgsimp} can be used to reduce the size of a well-formed $ADG$.

\algdef{SE}[DOWHILE]{Do}{doWhile}{\algorithmicdo}[1]{\algorithmicwhile\ #1}%
\begin{algorithm}[H]
\caption{\footnotesize Algorithm for reducing the size of a well-formed ADG built from a DT without considering redundant paths}\label{alg:wbadgsimp}
  \footnotesize
\begin{algorithmic}[1]
\Require A DT and a well-formed $ADG = \langle Ar, \mathscr{R}\rangle$ built from it
\Ensure An updated $ADG' = \langle Ar', \mathscr{R}'\rangle$
\State $Ar' \gets Ar$, $\mathscr{R}' \gets \mathscr{R}$
\Do
    \State $Ar'_{old} \gets Ar'$
    \ForEach {pair of arguments $ a = \langle f_i(v_{1}), y_m\rangle$ and $b = \langle f_i(v_{2}), y_m\rangle$}
    \Statex \Comment{\parbox[t]{0.93\linewidth}{\textit{Each pair of predictive arguments with same conclusion and support feature}}}
        \If{$f_i(v_{1}) \cup f_i(v_{2}) \neq \mathscr{V}_{f_i}$} \Comment{\parbox[t]{.5\linewidth}{\textit{if $v_{1}$ and $v_{2}$ do not cover all possible values of $f_i$, they do not need to be removed}}}
        \State \textbf{continue} 
        \EndIf
        
        \If{links $l_a$ and $l_b$ of $a$ and $b$ do not start from a same node $N$}
        \State \textbf{continue} \Comment{\parbox[t]{.7\linewidth}{\textit{ they do not need to be removed}}}
        \EndIf
        \Statex \Comment{\parbox[t]{0.89\linewidth}{\textit{Any pair of arguments at this point needs to be removed}}}
        \State $Ar' \gets Ar' \setminus \{a, b\}$ \Comment{\parbox[t]{.65\linewidth}{\textit{remove $a$ and $b$}}}
        \State $\mathscr{R}' \gets \mathscr{R}' \setminus (x,y)$ where $x \in {a,b}$ or $y \in {a,b}$ \Comment{\parbox[t]{.33\linewidth}{\textit{remove attacks with $a$ or $b$}}}
        \ForEach {$ c = \langle f(v), \emptyset\rangle \in Ar'$} \Comment{\parbox[t]{.40\linewidth}{\textit{for each non predictive argument}}}
            \If{link $l_c$ starts at $M$ and ends at $N$}
                \State $c \gets \langle f(v), y_m\rangle$ \Comment{\parbox[t]{.60\linewidth}{\textit{Add same conclusion as $a$ and $b$}}}
                \State \textbf{break}
            \EndIf
        \EndFor
    \EndFor
\doWhile{$Ar'_{old} \neq Ar'$} \Comment{\parbox[t]{.7\linewidth}{\textit{Repeat while there are arguments to remove}}}
\\
\Return $ADG' = \langle Ar', \mathscr{R}'\rangle$
\end{algorithmic}
\end{algorithm}

\begin{figure}
    \centering
\begin{tikzpicture}
\node[inner sep=0pt] (diagram) at (0,0)
    {\includegraphics[width=0.95\textwidth]{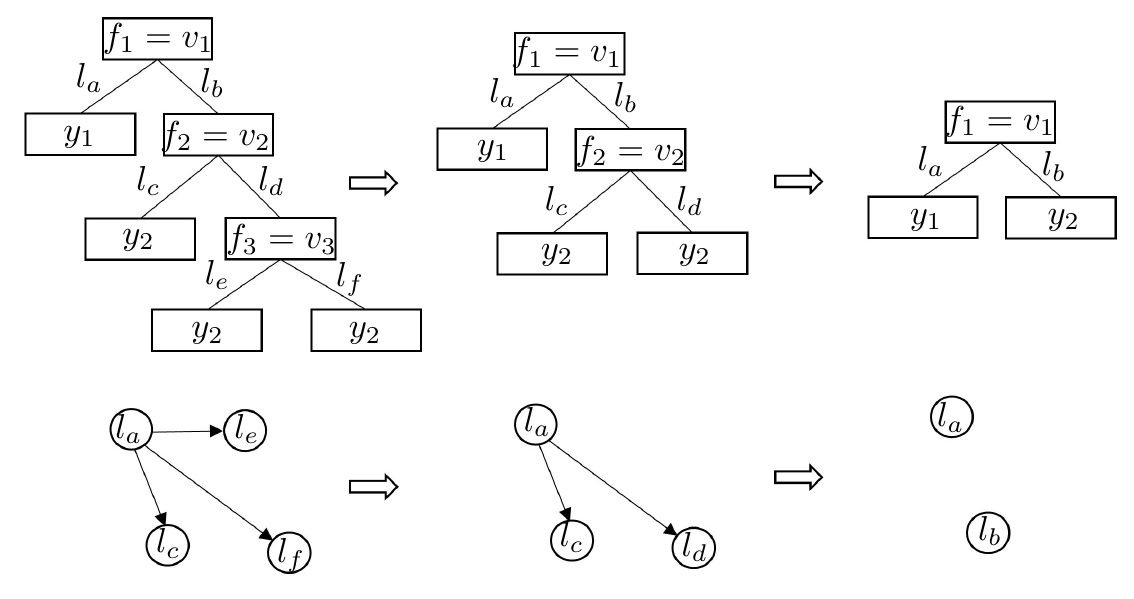}};
\end{tikzpicture}
\caption{Example of two steps of Algorithm \ref{alg:wbadgsimp} being employed to simplify an $ADG$. Each DT is depicted on top, with the corresponding $ADG$ below. Links between nodes are used to identify the arguments in the $ADG$.}
    \label{fig:simplification}
\end{figure}

The algorithm is designed to remove predictive arguments with the same conclusion that originate from a same node $N$ in the DT. Since the arguments will be removed, any attacks that make use of them also need to be removed. Finally, the non predictive argument originated from a link of another node $M$ to $N$ needs to have its conclusion updated. This process is repeated while there are arguments to be removed. Figure \ref{fig:simplification} depicts an example.

 \subsection{Extended Argumentation Graphs from Data (xADG)}

In order to allow for the use of more features by each argument's support, a $xADG$ in introduced here. Basically, a $xADG$ is an $ADG$ where each argument has an internal structure composed of boolean logical operators and one or more supports. It is defined as following:

\begin{definition}\label{def:xadg}
An extended argumentative decision graph \textbf{xADG}, is an argumentation framework $AF =
\langle Ar, \mathscr{R} \rangle$ where each $a \in Ar$ is defined as $a = \langle$($\phi_1$ \textsc{AND} $\phi_2$)
\textsc{OR} ($\phi_3$ \textsc{AND} $\phi_4$), $\theta$$\rangle$, $\phi_i \in \mathscr{P}_f, \theta \in \mathscr{P}_y \cup \varnothing$, and $\mathscr{R} \subseteq Ar \times Ar$.
\end{definition}

Def. \ref{def:xadg} characterises an argument in a $xADG$ without compromising its generality. This means that an argument can have a support with multiple predicates $\phi \in \mathscr{P}_f$ in any combination of boolean logical operators AND and OR. Furthermore, an $ADG$ is a specific case of a $xADG$ where all arguments' supports contain only one predicate $\phi \in \mathscr{P}_f$. The notion of well-built arguments in a $xADG$ is also introduced:

\begin{definition}\label{def:wbarg}
Given an xADG represented as an argumentation framework $AF =
(Ar, \mathscr{R})$, an argument $a \in Ar$ with $a = \langle$($\phi_1$ \textsc{AND} $\phi_2$)
\textsc{OR} ($\phi_3$ \textsc{AND} $\phi_4$), $\theta$$\rangle$ and $\phi_i = f_i(v_i)$ is said to be \textbf{well-built} iff it holds that:

\begin{enumerate}
    \item $f_1 \neq f_2$ and $f_3 \neq f_4$.
    \item One of the two options hold:
    \begin{enumerate}
        \item $f_{i} \neq f_{j}$, where $i \in \{1,2\}$ and $j \in\{3,4\}$, or
        \item $f_{i} = f_{j}$ and $v_i \neq v_j$, where $i \in {1,2}$ and $j \in{3,4}$
    \end{enumerate}
    
\end{enumerate}

\end{definition}

Informally, the first constraint guarantees that no same features are concatenated by the AND operator, while the second constraint guarantees that there is no redundancy by different clauses concatenated by the OR operators.

\subsection{Building xADGs from ADGs}

\begin{figure}[h]
    \centering
\begin{tikzpicture}
\node[inner sep=0pt] (diagram) at (0,0)
    {\includegraphics[width=0.70\textwidth]{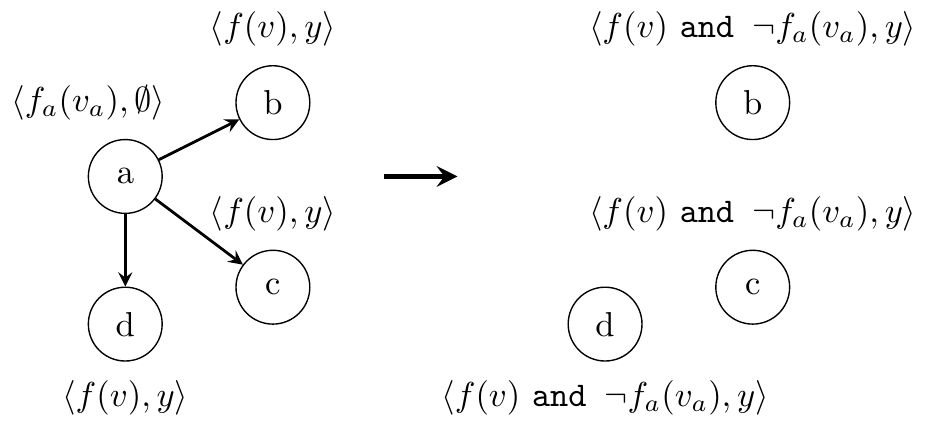}};
\node[inner sep=0pt] (diagram) at (0,2.4) {\textbf{Modification $\bm{m_1}$}};
\end{tikzpicture}
\caption{Example of modification ($m_1$) for removing non predictive arguments. The support of a non predictive attacking argument ($a$) is appended in the negative form to the support of the attacked arguments ($b$, $c$, and $d$) using the AND operator. $f(v)$ and $y$ are used to represent any feature with certain value and any conclusion. $f_a(v_a)$ is used to represent a specific feature and value used by argument $a$.}
    \label{fig:simplification1}
\end{figure}

\begin{figure}[h]
    \centering
\begin{tikzpicture}
\node[inner sep=0pt] (diagram) at (0,0.2)
    {\includegraphics[width=1\textwidth]{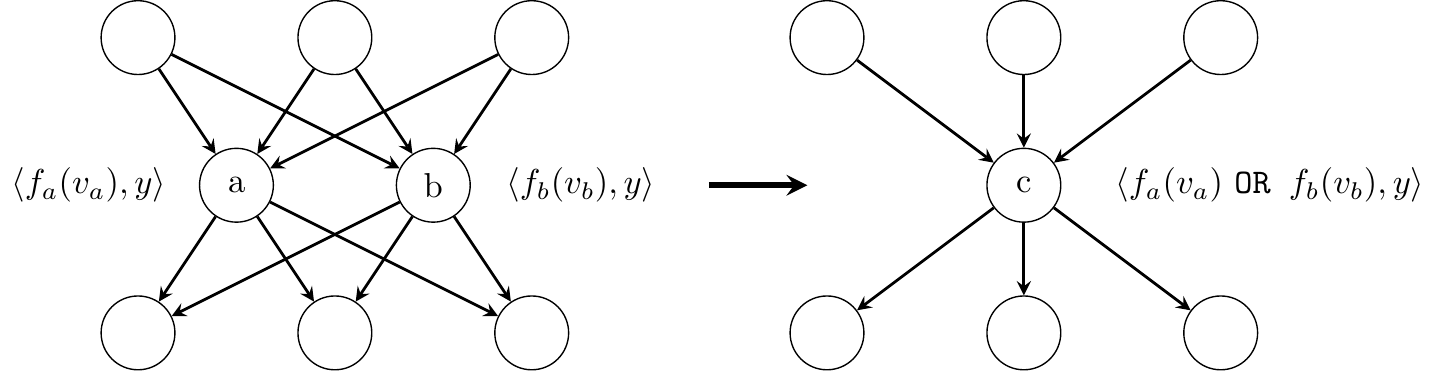}};
    \node[inner sep=0pt] (diagram) at (0,2.4) {\textbf{Modification $\bm{m_2}$}};
\end{tikzpicture}
\caption{Example of modification ($m_2$) for merging two predictive arguments. Two predictive arguments, $a$ and $b$, with the same conclusion and same set of targets and attackers (including empty sets) can be merged in a single argument. Supports are concatenated by the OR operator and the conclusion is kept the same.}
    \label{fig:simplification2}
\end{figure}

\begin{figure}[h]
    \centering
\begin{tikzpicture}
\node[inner sep=0pt] (diagram) at (0,0.7)
    {\includegraphics[width=1\textwidth]{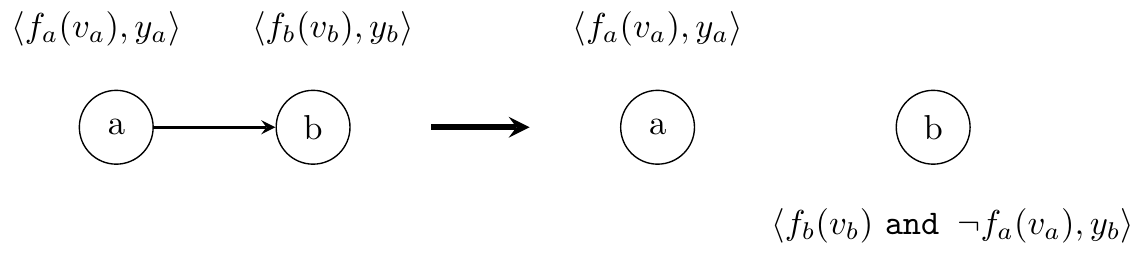}};
    \node[inner sep=0pt] (diagram) at (0,2.4) {\textbf{Modification $\bm{m_3}$}};
\end{tikzpicture}
\caption{Example of modification ($m_3$) for removing an attack. The attack from $a$ to $b$ can be removed by appending the negative support of the attacker to the target argument using the AND operator.}
    \label{fig:simplification3}
\end{figure}

Given a well-formed $ADG$, a $xADG$ of equivalent inferential capability can be built by performing a set of modifications. In other words, attacks and arguments can be reduced in a $ADG$ by modifying the internal structure of arguments, as defined by an $xADG$. This can be done while keeping the same set of inferences produced by the original well-formed $ADG$. Figures \ref{fig:simplification1}-\ref{fig:simplification3} introduce three of such modifications, named $m_1, m_2$ and $m_3$. 

Concerning the equivalence of inferences, when applying $m_1$ or $m_3$, the attacked arguments that were initially rejected (accepted) from some extension-based semantics, will now have their supports in the resulting $xADG$ evaluate to false (true). Hence, the inferences of predictive arguments are preserved. As for $m_2$, instead of having one or two arguments initially activated, only one in the resulting $xADG$ will be activated (with the same conclusion). Since the attacks of the merged argument remain unchanged, it will be accepted or rejected in the same way as any of the other two in the original structure\footnote{This characteristic does not hold for ranking-based semantics \cite{Amgoud2013ranking}, in which the number of attackers and attacked arguments may affect the arguments' acceptance.}. 

Regarding the amount of supports resulting from $m_{1-3}$, it is worth noting that the examples used to introduce them had initially arguments with only one support, leading to a $xADG$ with arguments having two supports. These simplified examples are easier to comprehend, but it is possible that the changes performed by $m_{1-3}$ may result in a larger number of supports. For instance, applying $m_2$ repeatedly, can lead to an argument having several OR clauses. Hence, the potential advantages and disadvantages of these modifications in terms of number of attacks/arguments and amount of supports is detailed below:

\begin{itemize}[topsep=0pt,itemsep=1.0ex,partopsep=1ex,parsep=1ex]
    \item[\textbullet] Modification $m_1$: this has the potential to remove a single argument and multiple attacks originated from it, by concatenating its negated support to the attacked arguments using the AND operator. When arguments are well-built, it is possible that the number of supports in the resulting argument will not increase.
    \item[\textbullet] Modification $m_2$: this has the potential to remove a single argument and all the attacks originated and targeted at it. The cost is the addition of its support/s using the OR operator to a second argument with the same conclusion and set of attackers and targets. If arguments are well-built, it is possible that the number of supports in the resulting argument is less than the sum of the supports in the two original arguments. The trade-off between number of arguments/attacks and number of supports will depend on how many attacks can be remove, the amount of supports in the resulting argument, and whether the arguments are enforced to be well-built.
    \item[\textbullet] Modification $m_3$: this has the potential to reduce a single attack by adding its negative support/s to the attacker using the AND operator. 
\end{itemize}

In terms of enhancing explainability, $m_1$ and $m_2$ seem to be more promising, since they have the potential to remove multiple attacks and arguments. $m_3$ seems to be less promising in terms of size of the resulting $xADG$ and the amount of arguments' supports. The exchange when using $m_3$ is for one less attack and at least one more support in the attacked argument. Moreover, it can be argued that the visual representation of an attack is a better way of depicting contradictions between arguments instead of negative supports.

\subsection{Experimental setup}

To investigate if small $xADGs$ can be achieved, an initial evaluation is proposed. First, a dataset and a respective DT need to be provided. Next, algorithms \ref{alg:wbadg} and \ref{alg:wbadgsimp} can be applied followed by a series of modifications $m_2$ and $m_1$. $m_2$ is applied before $m_1$, since merging arguments after removing attacks, could lead to more merged arguments with a high amount of supports. In turn, $m_3$ is not employed, since it does not offer a good trade-off in terms of explainability. A max number of supports can be set when using $m_1$ and $m_2$, so as to avoid the number of supports to be too high. However, this is not limited in this analysis in order to give a better notion of what can be achieved with $m_1$ and $m_2$. Algorithm \ref{alg:evaluation} details the steps of this evaluation.

\algdef{SE}[SUBALG]{Indent}{EndIndent}{}{\algorithmicend\ }%
\algtext*{Indent}
\algtext*{EndIndent}

\begin{algorithm}[h]
\caption{\footnotesize Proposed algorithm for the creation of $xADGs$ built from DTs}\label{alg:evaluation}
\footnotesize
\begin{algorithmic}[1]
\Require Dataset $\mathscr{D}$
\Ensure a $xADG$

\State \textbf{Function} WellBuiltArguments($xADG = \langle Ar, \mathscr{R}\rangle$):
\Indent
\ForEach {$a \in Ar$}
\State $a \gets$ wellBuilt($a$) \Comment{\parbox[t]{.6\linewidth}{\textit{Updates $a$ by removing any redundancy and making sure it is well-built}}}
\EndFor
\\
\Return $xADG$
\EndIndent
\State $DT \gets $ A DT classifier$(\mathscr{D})$ \Comment{\parbox[t]{0.55\linewidth}{\textit{Some DT algorithm, such as C4.5 or CART}}}

\State $ADG \gets$ Algorithm \ref{alg:wbadg} ($DT$)
\State $ADG \gets $  Algorithm \ref{alg:wbadgsimp} ($DT, ADG$)

\State $xADG \gets ADG$ \Comment{\parbox[t]{.7\linewidth}{\textit{First version of $xADG$ before applying $m_1$ and $m_2$}}}

\Do
    \State $xADG_{old} \gets xADG$
    \Statex \Comment{\parbox[t]{.932\linewidth}{\textit{Merge any single pair of arguments and update it to be well-built}}}
    \State $xADG \gets $ WellBuiltArguments($m_2(xADG)$)
\doWhile{$xADG_{old} \neq xADG$} \Comment{\parbox[t]{.6\linewidth}{\textit{Repeat while there are arguments to merge}}}

\Do
    \State $xADG_{old} \gets xADG$
    \Statex \Comment{\parbox[t]{.932\linewidth}{\textit{Remove any single attacking argument, add its negative support to attacked arguments, and update attacked arguments to be well-built}}}
    \State $xADG \gets $ WellBuiltArguments($m_1(xADG)$)
\doWhile{$xADG_{old} \neq xADG$} \Comment{\parbox[t]{.6\linewidth}{\textit{Repeat while there are arguments arguments and attacks to remove}}}
\\
\Return $xADG$
\end{algorithmic}
\end{algorithm}

\section{Results}

In order to assess if small $xADGs$ can be achieved for problems of different magnitudes, four datasets from the UCI machine learning repository \cite{uci-ml} were chosen\footnote{\texttt{archive.ics.uci.edu/ml/datasets/Car+Evaluation} \texttt{archive.ics.uci.edu/ml/datasets/Adult} 
\\
\texttt{archive.ics.uci.edu/ml/datasets/Bank+Marketing}
\texttt{archive.ics.uci.edu/ml/datasets/Myocardial+infarction+complications}}. These datasets were selected based on their number of features and records, so as to provide a more robust indication of the usefulness of $xADGs$. A DT was built for each one with the optimised version of the CART algorithm provided by the \texttt{scikit-learn} package \cite{scikit-learn}. Training and testing sets used a 80\%-20\% split ratio. The maximum depths of the DTs were chosen based on a preliminary analysis of balanced accuracy. An increasing number of maximum depths were evaluated, and the last one to offer a reasonable improvement in balanced accuracy was chosen. This was done to achieve the highest balanced accuracy without growing the tree for irrelevant gains. Table \ref{tab:results} lists the results for the 4 datasets. The accuracy, balanced accuracy, tree size (nodes and edges), average path length, $xADG$ size (arguments and attacks), and number of supports (min, average and max) are reported. The CI 95\% is presented for all results after running a number of 100 executions for each dataset. The work in \cite{dondio2021towards} performs a similar experiment for building well-formed $ADGs$ using the cars and census dataset. A greedy algorithm is proposed to build the $ADG$ from scratch. Table \ref{tab:resultscomparison} lists a comparison with this approach.

\subsection{Analysis and discussion}

In all experiments, as reported in Table \ref{tab:results}, the resulting $xADGs$ achieved satisfactory size. That means they are considered small enough to be reasonably understandable by human reasoners. This assumption might not hold if a higher number of supports is employed. However, this should be evaluated with user studies, and it is suggested as future work.

\setlength{\tabcolsep}{0.0pt}
\renewcommand{\arraystretch}{1.3} 
\footnotesize
\begin{table}[h]
\caption{Summary statistics of DTs and $xADGs$ found for 4 different datasets. Ranges represent the CI 95\%. For the cars dataset, class values \textit{acc}, \textit{good}, \textit{vgood} were grouped into \textit{acc}. For the myocardial dataset, ZSN was chosen as target feature since it is better balanced, and features with more than 5\% missing data were dropped.
}\label{tab:results}
\begin{tabular}{{P{3.5cm}p{2.18cm}p{2.18cm}p{2.18cm}p{2.18cm}}}
\hline
& \multicolumn{4}{c}{\textbf{Datasets}}
\\
\cline{2-5}
& \textbf{Cars} & \textbf{Census} & \textbf{Bank} & \textbf{Myocardial}
\\
\cline{2-5}
\textbf{Features} & 6 & 14 & 16 & 59
\\
\textbf{Records} & 1728 & 30152 & 45211  & 1436
\\
\cline{2-5}
\textbf{Tree depth} & 6 & 4 & 3 & 3
\\
\textbf{Avg. path length} & [5.1, 5.2] & [3.8, 3.9] & [3.0, 3.0] & [3.0, 3.0]
\\
\textbf{Tree nodes} & [26.6, 27.3] & [26.7, 27.4] & [15, 15] & [15,15]
\\
\textbf{Tree edges} & [25.6, 26.3] & [25.7, 26.4] & [14, 14] & [14,14]
\\
\cline{2-5}
\textbf{$xADG$ args.} & [8.3, 9.0] & [8.4, 8.9] & [6.8, 7.0] & [3.2, 3.7]
\\
\textbf{$xADG$ atts.} & [7.7, 9.6] & [6.4, 8.3] & [2.0, 2.0] & [1.0, 1.3]
\\
\textbf{Supports min} & [1.8, 2.0] & [2.0, 2.1] & [2.0, 2.0] & [1.1, 1.3]
\\
\textbf{Supports max} & [3.3, 3.6] & [3.9, 4.0] & [2.9, 3.0] & [1.2, 1.4]
\\
\textbf{Supports average} & [2.6, 2.7] & [2.5, 2.6] & [2.3, 2.3] & [1.1, 1.3]
\\
\cline{2-5}
\textbf{Accuracy} & [0.940, 0.946] & [0.839, 0.841] & [0.898, 0.900] & [0.796, 0.809]
\\
\textbf{Balanced Acc.} & [0.939, 0.946] & [0.728, 0.732] & [0.651, 0.658] & [0.600, 0.611]
\\
\hline
\end{tabular}
\end{table}
\normalsize
\setlength{\tabcolsep}{0.0pt}

\renewcommand{\arraystretch}{1.3} 
\footnotesize
\begin{table}[H]
\caption{Comparison of well-formed $ADGs$ and $xADGs$ found for the cars and census dataset. Ranges represent the CI 95\%.}\label{tab:resultscomparison}
\begin{tabular}{{P{3.5cm}p{2.18cm}p{2.18cm}p{2.18cm}p{2.18cm}}}
\hline
& \multicolumn{4}{c}{\textbf{Datasets}}
\\
\cline{2-5}
& \textbf{Cars \cite{dondio2021towards}} & \textbf{Cars} & \textbf{Census \cite{dondio2021towards}} & \textbf{Census}
\\
\cline{2-5}
\textbf{Accuracy} & 0.88 & [0.940, 0.946] & 0.83 & [0.839, 0.841]
\\
\textbf{Balanced Acc.} & 0.88 & [0.939, 0.946] & 0.75 & [0.728, 0.732]
\\
\textbf{Arguments} & 8 & [8.3, 9.0] & 21 & [8.4, 8.8]
\\
\textbf{Attacks} & 11 & [7.7, 9.0] & 78 & [6.4, 8.3]
\\
\textbf{Supports min} & 1 & [1.8, 2.0] & 1 & [2.0, 2.1]
\\
\textbf{Supports max} & 1 & [3.3, 3.6] & 1 & [3.9, 4.0]
\\
\textbf{Supports average} & 1 & [2.6, 2.7] & 1 & [2.5, 2.6]
\\
\hline
\end{tabular}
\end{table}
\normalsize

When comparing the $xADG$ with their counterpart DTs, it is essential to consider both the number of supports and the average path length in the DT. Although the average number of supports may be lower, it is important to note that attacks in the $xADG$ might require multiple arguments (if all activated) to produce an inference. Contrarily, the nodes in a DT path can always produce an inference without interacting with the nodes in another path. Therefore, the number of attacks produced should also be considered. This was typically smaller or substantially smaller than the number of arguments reported, suggesting that the evaluation of attacks between arguments might not be necessary to produce an inference in many cases.
Hence, when using model size as a metric for assessing users' understandability of both the model and the inferences it generates, there seem to be pros and cons for the DT and for the equivalent $xADG$. The results provided by this quantitative analysis do not seem to offer a decisive outcome. Depending on the available data and prior knowledge of the human reasoner, either the conflict-based representation of $xADG$ or the mutually exclusive rules generated by the paths in the DT could result in a model that is more concise and understandable. However, it is crucial to observe that the $xADG$ allows knowledge to be more easily incorporated, through the addition of arguments and attacks, without any restriction in its topology. Thus, it can be argued that it is more adequate for knowledge discovery, acquisition and refinement.

As for the comparison in Table \ref{tab:resultscomparison}, note that the accuracy and balanced accuracy are better for the $xADGs$ in the cars dataset, and similar in the census dataset. However, the number of arguments and attacks is smaller for the cars dataset and substantially smaller for the census dataset. The number of supports is kept small enough for the arguments to be assumed understandable as previously discussed. This suggests that leveraging the structure and inferential capability of DTs with $xADGs$ might be a good alternative for automating the creation of structured argumentation frameworks with stronger (balanced) accuracy and reasonable size to be used by human reasoners.

\section{Conclusions}

In this paper, a novel framework for structured argumentation, named extend argumentative decision graph ($xADG$), was proposed. It is an extension of the argumentative decision graph ($ADG$) proposed by \cite{dondio2021towards} and itself an extension of Dung's abstract argumentation graphs. It enabled arguments to use boolean logic operators and multiple supports within their internal structure. Therefore, it is able to produce more concise argumentation graphs and potentially enhance its understandability from its users' point of view. Results for classification problems of different sizes indicate that xADGs can be built with strong predictive capacity, derived from an input decision tree, while potentially lowering the average number of premises to reach a conclusion. Comparisons with other techniques for construction of $ADGs$ also indicate a better predictive capacity coupled with smaller model size. It is expected that $xADGs$ will provide an initial step for the creation of more concise structured argumentation frameworks that can be learned from data and used for knowledge discovery, acquisition and refinement.

Future work can consider the use of more optimised DTs \cite{kotsiantis2013decision} that could lead to smaller $xADGs$, as well as $xADGs$ modifications that can impact on the models' inferential capability. It is possible that modifications that slightly reduce predictive capacity might allow significant gains in understandability. It is also important that user studies are performed to evaluate the understandability of $xADGs$ with a larger number of supports originated from massive decision trees. The addition of human knowledge by domain experts to $xADGs$ could also be tested in various contexts. Finally, $xADGs$ can be employed to reduce the size of large $ADGs$ not linked to monotonic, data-driven models (as in \cite{dondio2021towards}), allowing a more diverse range of understandable solutions to be found.

\bibliographystyle{splncs04}
\bibliography{references}
%




\end{document}